\documentclass{article} 
\usepackage{iclr2024_conference,times}

\usepackage{amsmath,amsfonts,bm}

\def\eqref#1{equation~\ref{#1}}

\def\1{\bm{1}}

\DeclareMathAlphabet{\mathsfit}{\encodingdefault}{\sfdefault}{m}{sl}
\SetMathAlphabet{\mathsfit}{bold}{\encodingdefault}{\sfdefault}{bx}{n}

\usepackage{hyperref}
\usepackage{url}
\usepackage{graphicx}
\usepackage{enumitem}
\usepackage{multirow}
\usepackage{amsmath}
\usepackage{xcolor,colortbl}
\usepackage{soul}
\usepackage{wrapfig}

\usepackage{amssymb}
\usepackage{pifont}

\newcommand{\xmark}{\ding{55}}

\def\ours{\texttt{bio}FAME~}
\colorlet{FA-let}{white!85!cyan}
\colorlet{FM-let}{white!85!gray}
\definecolor{FA}{RGB}{55, 172, 226}
\definecolor{FM}{RGB}{190, 190, 190}

\title{Frequency-Aware Masked Autoencoders\\ for Multimodal Pretraining on Biosignals}

\author{Ran Liu$^{1, 2}$\thanks{Work completed during internship at Apple. Contact: rliu361@gatech.edu, amoin@apple.com.} , Ellen L. Zippi$^{1}$, Hadi Pouransari$^{1}$, Chris Sandino$^{1}$, Jingping Nie$^{1, 3}$$^*$ \\ \textbf{Hanlin Goh$^{1}$, Erdrin Azemi$^{1}$, Ali Moin$^{1}$}  \\ 
Apple$^1$, Georgia Institute of Technology$^2$, Columbia University$^3$
}

 \iclrfinalcopy 

\begin{document}




\maketitle

\begin{abstract}
Leveraging multimodal information from biosignals is vital for building a comprehensive representation of people's physical and mental states. 
However, multimodal biosignals often exhibit substantial distributional shifts between pretraining and inference datasets, stemming from changes in task specification  or variations in modality compositions.
To achieve effective pretraining in the presence of potential distributional shifts,
we propose a frequency-aware masked autoencoder (\ours \hspace{-1mm}) that learns to parameterize the representation of biosignals in the frequency space.
\ours incorporates a frequency-aware transformer, which leverages a fixed-size Fourier-based operator for global token mixing, 
independent of the length and sampling rate of inputs.
To maintain the frequency components within each input channel, we further employ a frequency-maintain pretraining strategy that performs masked autoencoding in the latent space.
The resulting architecture effectively utilizes multimodal information during pretraining, and can be seamlessly adapted to diverse tasks and modalities at test time, regardless of input size and order.
We evaluated our approach on a diverse set of transfer experiments on unimodal time series, achieving an average of $\uparrow$$5.5\%$ improvement in classification accuracy over the previous state-of-the-art. Furthermore, we demonstrated that our architecture is robust in modality mismatch scenarios, including unpredicted modality dropout or substitution, proving its practical utility in real-world applications. Code is available at \href{https://github.com/apple/ml-famae}{https://github.com/apple/ml-famae}.
\end{abstract}

\section{Introduction}
Physical and mental states of an individual are 
manifested by a variety of physiological responses or \textit{biosignals}. 
For example, 
electroencephalography (EEG) can decode human emotions by monitoring their brain activities \citep{liu2010real}, electromyography (EMG) can detect facial expressions such as smiling by recording muscle contractions \citep{canento2011multimodal}, 
and a combination of these modalities can help decode a person's affective states.
The effective use of multimodal information can not only build better and more resilient representations of the human body and mental states \citep{bachmann2022multimae,smith2005development,de1998category}, but also help researchers understand how each biosignal contributes to each physiological state
and how their information overlaps \citep{bird2020cross}.

\begin{figure}[t]
\begin{center}
\includegraphics[width=0.9\textwidth]{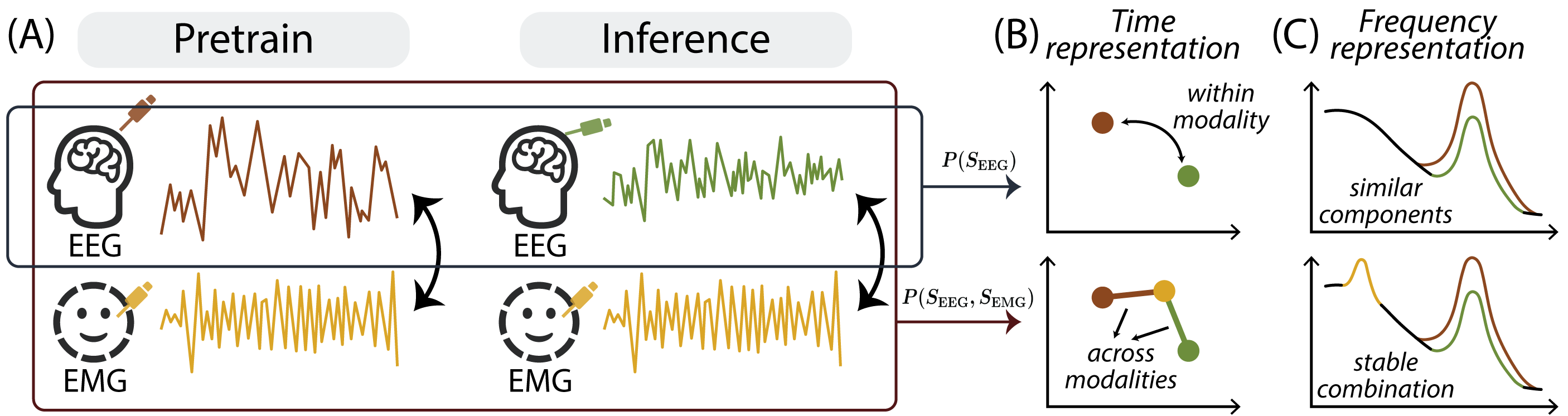}
\caption{\small \textit{Motivation of our approach}. \textbf{(A)} In multimodal biosignal systems, there exists substantial distributional shifts between the pretraining and inference datasets. \textbf{(B)} The distributional shifts often cause the shifts of representation in time-space, which can affect the model's generalization ability within modality and across modalities. \textbf{(C)} In the meantime, the representation in frequency-space typically would contain similar frequency components within modality, leading to more stable combinations in multimodal scenarios.} 
\label{fig:motivation}
\vspace{-4mm}
\end{center}
\end{figure}

Recently, in language-vision domains, large-scale multimodal pretraining has demonstrated remarkable generalization and zero-shot capabilities \citep{huang2021makes,bachmann2022multimae,radford2021learning}, outperforming small-scale models that are trained on specific downstream tasks \citep{kirkpatrick2017overcoming,radford2019language}.
In light of these advancements, we investigate whether similar pretraining can be applied to the biosignal domain.
However, performing multimodal pretraining on biosignals is particularly challenging due to the significant distributional shifts between the pretraining and downstream datasets.
This challenge can be categorized in two ways:
(i) For biosignals, there are substantial distributional \textbf{shifts within each modality}, wherein 
data varies across tasks, subjects, and even recording sessions within subjects due to slight changes in sensor placement and recording conditions \citep{cheng2020subject}. Additionally,
(ii) multimodal biosignals might encounter strong distributional \textbf{shifts across modalities}, 
meaning that the connection between different modalities can be altered.
These crossmodal domain shifts can arise from unimodal shifts, as a change in a single modality can disrupt its relationship to a different modality. 
Moreover, multimodal biosignals often face \textit{modality mismatch scenarios}, 
where modalities may be unavailable at test time, and thus are removed or replaced with new modalities that provide relevant information to the detected physiological response \citep{mckinzie2023robustness}.
Addressing these distributional shifts is crucial to effectively leverage multimodal pretraining on biosignals.

In this work, we propose to incorporate frequency information in time series to mitigate distributional shifts and enable multimodal pretraining on biosignals.
Frequency-domain analysis is advantageous for biosignals not only due to its invariance to common causes of distributional shifts such as temporal shifts and scaling, but also because the extracted frequency components are characteristic representations for physiological activities (see Figure~\ref{fig:motivation}).
While previous works have shown the effectiveness of using frequency domain information to address generalization issues, they have either relied on encoders from both the time and frequency domains \citep{zhang2022self}, or complicated sampling and combining modules \citep{zhou2022fedformer} to utilize the frequency information. Here, we propose a simple, yet effective, multi-head frequency filter layer
with fixed-size Fourier-based operator
that directly parameterizes the representation of biosignals in the frequency space.
The proposed layer can be easily incorporated into the transformer, giving a \textit{frequency-aware (FA) encoder} that is both expressive and computationally efficient.

Furthermore, to extend the frequency awareness into a multimodal pretraining setting, we couple the FA encoder with a \textit{frequency-maintain (FM) pretraining strategy}.
To prevent the statistical consistency within the data from being disrupted by conventional masked autoencoding strategies \citep{ryali2023hiera}, our method performs masked autoencoding in the latent space to maintain the frequency awareness during reconstruction.
Coupled with a channel-independent design \citep{nie2022time,liu2022seeing},
our model presents a pure reconstruction-based multimodal pretraining architecture that can effectively combine and utilize information across 
modalities, 
with robustness towards distributional shifts within and across modalities.

To systematically evaluate our proposed approach, 
we first examine the transferability of our architecture on a publicly available one-to-many transfer learning benchmark \citep{zhang2022self}.
Our architecture achieves state-of-the-art performance, giving an average of $\uparrow$$5.5\%$ improvements in classification accuracy over the previous state-of-the-art, showing consistency 
across datasets of different input lengths, sampling rates, and diverse sources of modalities. 
Next, we demonstrate that our architecture is robust to a variety of modality mismatch scenarios commonly encountered in real-world cases, showing that 
our architecture can effectively integrate and leverage information across multiple modalities during pretraining.

We summarize our main contributions as follows:
\begin{itemize}[]
\itemsep0em 
    \item We propose \ours\hspace{-1mm}, a frequency-aware masked autoencoder for biosignals comprising: (i) a frequency-aware (FA) transformer encoder that can learn biosignals in a robust and computationally efficient way; (ii) a frequency-maintain (FM) pretraining strategy that retains the frequency awareness during reconstruction.
    \item By constructing a fixed-size Fourier-based operator in the architecture, \ours can be {pretrained on multimodal biosignals and adapted to new modalities} of varying lengths and frequency components, exhibiting resilience to distributional shifts even when the modalities differ between training and testing.
    \item \ours achieves consistently robust performance on a diverse set of transfer experiments, outperforming the previous state-of-the-art by an average improvement of $\uparrow$$5.5\%$, demonstrating how utilizing multimodal information at the pretraining stage can benefit the generalization ability of the model.
\end{itemize}

\section{Background}
\label{sec:background}

\paragraph{Multimodal Pretraining Methods}
Pretraining large-scale models that can effectively use multimodal information has gathered a lot of research attention due to its strong capability of generalization \citep{huang2021makes,liang2022foundations,reed2022generalist,chai2022deep}.
Multimodal pretraining methods can be roughly categorized as 
(i) those that train separate encoders for each modality, as seen with contrastive methods that design novel objectives to align or fuse  representations from different modalities \citep{li2021align,radford2021learning,jia2021scaling}, 
and (ii) those that design one unified architecture for many modalities, with completely shared encoders per-modality or a few layers shared for decoding \citep{reed2022generalist,akbari2021vatt,wang2022ofa}. 
The benefit of using one unified architecture is that we can build a joint representation space that connects different modalities, as well as share weights
to reduce additional computational overhead \citep{bachmann2022multimae,lu2022unified}.
Inspired by the latter, our work aims to train a single unified architecture for multimodal biosignals with an effective frequency-awareness design.

\paragraph{Pretraining on Biosignals and Time Series}

Biosignals are multivariate time series that capture various physiological processes within the human body \citep{giannakakis2019review,cheng2020subject}.
While biosignals are crucial for diverse applications such as human-computer interaction, 
acquiring an ample amount of labeled biosignals is a labor-intensive process that requires the involvement of domain experts \citep{ericsson2022self}.
To alleviate the need for labeled data, researchers proposed various self-supervised methods to pretrain the model with large-scale unlabeled datasets. This includes (i) contrastive methods that build latent representation based on similarity across samples of different augmentation \citep{cheng2020subject,kiyasseh2021clocs,zhang2022self}, (ii) reconstruction-based methods that perform either feature reconstruction or data reconstruction \citep{kostas2021bendr,chien2022maeeg}, or (iii) a hybrid of both \citep{dong2023simmtm}.
While previous works demonstrate that pretraining on large-scale data can benefit downstream task performance, 
however, most of the existing works only explored unimodal pretraining without investigating how to effectively utilize the multimodal information present at training time.
Existing work even shows that pretraining on multimodal information could cause performance degradation due to 
the large variation across modalities \citep{zhang2022self}. 
To the best of our knowledge, this is the first work that explores how to effectively perform multimodal pretraining on biosignals that gives robust performance towards distributional shifts within and across modalities.

\section{Motivation of Our Approach}
Parameterizing representations in the frequency space is shown to be effective in many domains.
Frequency-based approaches are particularly effective in solving partial differential equations and modeling long sequences \citep{li2020fourier,gu2021efficiently,li2022makes,zhou2022film}, as it can effectively capture long-range dependencies. 
{Frequency-aware approaches are also widely used in computer vision, as it can improve image fidelity and can effectively mix tokens when used in the transformer architecture \citep{rao2021global,guibas2021adaptive,xie2022masked,liu2022devil,li2022architecture}.
Akin to physiological signal processing, frequency-based approaches are employed to effectively extract discriminative patterns within sensory signals \citep{yao2019stfnets,li2021units}.}
The robustness of frequency-based operations can be partially attributed to the connection between Fourier transform and global circular convolution \citep{zhi2016generalized,li2020falcon}. 

Recently, many works suggest that the periodic oscillations and analogous patterns in the frequency space exhibit rich information for electrophysiological signals \citep{donoghue2020parameterizing,bird2020cross,subha2010eeg,demanuele2007distinguishing}.
Thus, several frequency-aware approaches are proposed to study biosignals.
For example, \cite{zhang2022self} used the consistency between time and frequency spaces to guide the learning on biosignals, demonstrating improved transferability and generalizability on downstream tasks. Other works perform cross-domain reconstruction across the time and spectral domains \citep{zhang2022cross,yang2022unsupervised}.

{
Contrary to prior studies, \ours emphasizes transferability and efficient adaptation to downstream tasks across many physiological modalities, by leveraging frequency-space information during pretraining on multimodal data to forge a universal representation of biosignals. 
We design novel mechanism and architecture to build a fully transferable and computation-efficient approach for frequency-aware representation extraction, 
setting \ours apart from conventional methods that are constrained by frequency-space encoders or decoding components tailored to specific input sizes \citep{wu2022neuro2vec}. 
These conventional methods often struggle with modality transfer due to varying frequency components and introduce unnecessary computational burdens and overparameterization. Our approach, in contrast, ensures flexibility and efficiency, free from such limitations.}




\begin{figure}[t]
\begin{center}
\includegraphics[width=\textwidth]{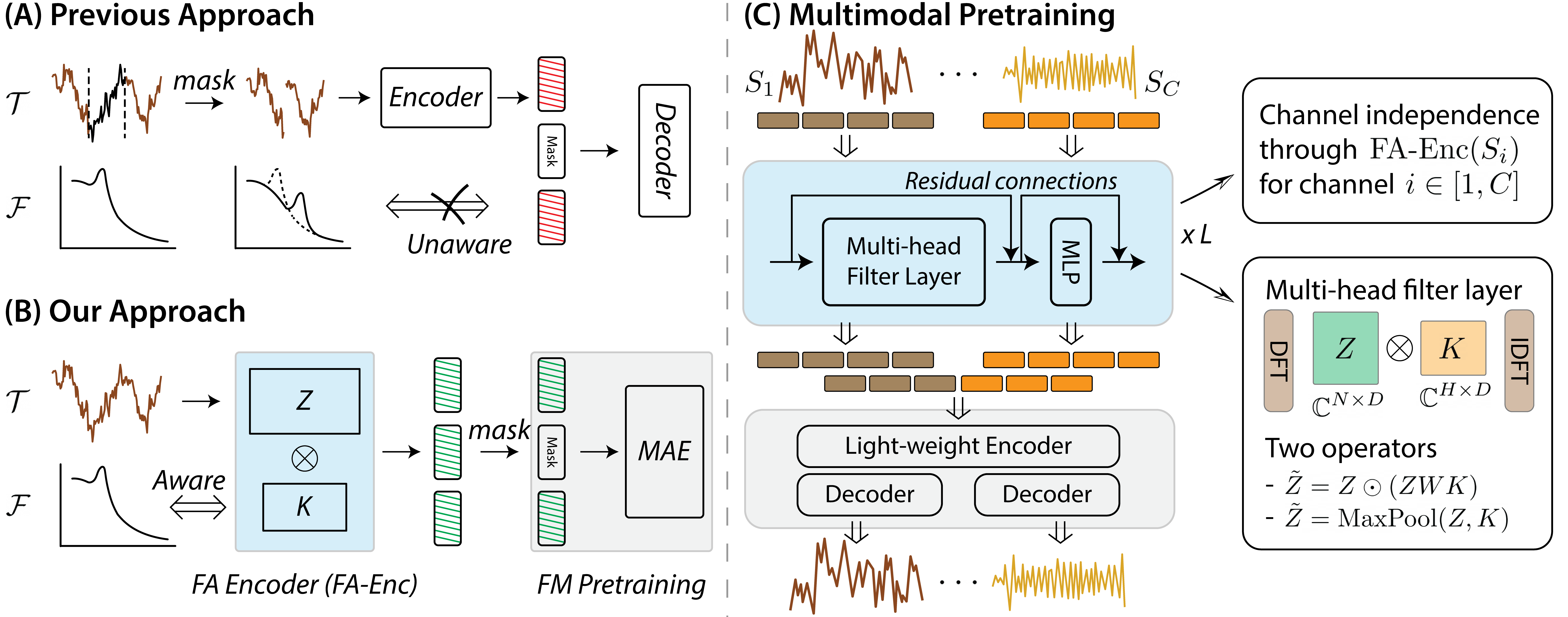}
\caption{\small \textit{Overview}. \textbf{(A)} Previous approaches perform masking in the time domain, which causes shifts in the frequency components. Also, the encoders are unaware of the frequency information in time series. \textbf{(B)} To address the issues, we propose \ours\hspace{-1mm}, which (i) builds frequency awareness by directly learning frequency filters in the representation space, and (ii) performs masked autoencoding in the latent space to maintain frequency information during pretraining. 
\sethlcolor{FA-let}
\textbf{(C)} We implement \ours in the multimodal pretraining scheme, where the \hl{frequency-aware encoder} ($\operatorname{FA-Enc}(\cdot)$) processes signals in a channel-independent manner, and extracts representations with multi-head filter layer with fixed-size Fourier operators. 
\sethlcolor{FM-let}
The \hl{frequency-maintain pretraining strategy} further performs masked autoencoding in the latent space with separate reconstruction to guide the effective mixing of multimodal information.} 
\label{fig:method}
\end{center}
\end{figure}

\section{Method}
\label{sec:method}


\paragraph{Preliminaries: Discrete Fourier Transform (DFT) for Token Mixing}

DFT is widely used in traditional methods for processing biosignals and images \citep{pitas2000digital}.
For a time space representation $\bm{x} \in \mathbb{R}^{N}$ with $N$ elements $x_n$, $n \in [0, N-1]$, its corresponding frequency space representation $\bm{z} \in \mathbb{C}^{N}$ with elements $z_k$ is  produced by DFT ($\mathcal{F}(\bm{x})=\bm{z}$), which can be inversed through the Inverse Discrete Fourier Transform (IDFT) ($\mathcal{F}^{-1}(\bm{z})=\bm{x}$) as below:
\begin{equation}
    \operatorname{DFT: } z_k=\sum\limits_{n=0}^{N-1} x_n e^{-i(2 \pi / N) k n}, \quad \operatorname{IDFT: } x_n = \frac{1}{N} \sum\limits_{k=0}^{N-1} z_k e^{i(2 \pi / N) k n},
\end{equation}
where $i$ is the imaginary unit.
The computational complexity of DFT can be reduced from quadratic to $\mathcal{O}(N \log N)$ when leveraging 
the fast Fourier transform (FFT) algorithm \citep{brigham1988fast}.

Consider a sequence $X = [\bm{x}_1, ..., \bm{x}_N]^T \in \mathbb{R}^{N \times D}$ of $N$ tokens of $D$-dimensions, transformers aim to learn the interactions across tokens, typically through the self attention operation.
Recently, mixing tokens with frequency-based operations through DFT and IDFT is shown to be a computationally efficient alternative \citep{rao2021global,guibas2021adaptive}, as it considers global-wise information mixing. 
The token mixing process is theoretically grounded by the Fourier Neural Operators \citep{li2020fourier}, which is often implemented in its discrete form (denote as $\mathcal{K}$) as such:
\begin{equation}
    (\mathcal{K}(X))(\bm{x}_i) = \mathcal{F}^{-1}(R \cdot \mathcal{F}(X))(\bm{x}_i), \forall i \in [1, N]
\end{equation}
Ideally, $R$ should be the Fourier transform of a periodic function which admits a Fourier series expansion. For the sake of simplicity, it is often implemented as learnable weights of shape $\mathbb{C}^{N \times D}$.

\subsection{Frequency-aware transformer with multi-head frequency filters}
\label{sec:methodFA}

In this work, we seek to understand two questions: (i) If parameterizing biosignals in the frequency space would provide better empirical performance, as frequency information is shown to be vital for many physiological activities; (ii) How to design a frequency-aware architecture that is transferrable and generalizable across different types of biosignals with varying input lengths and sampling rates.
To address those two questions, we propose a multi-head frequency filter layer to build a frequency-aware transformer encoder $\operatorname{FA-Enc(\cdot)}$.

\paragraph{Multi-head Frequency Filter Layer}
We propose to manipulate the frequency representation with a multi-head frequency filters $K \in \mathbb{C}^{H \times D}$, where $H$ is the total number of heads.
Given a sequence of tokens $X \in \mathbb{R}^{N \times D}$, 
we first perform DFT along the sequence dimension to obtain its representation in the frequency space as $Z \in \mathbb{C}^{N \times D}$.
To obtain the manipulated features in frequency space $\tilde{Z} \in \mathbb{C}^{N \times D}$, we first compute queries $Q = {Z}W$, where $W \in \mathbb{R}^{D \times H}$ is a learnable matrix that is used to combine processed information across different filters. The resulting queries are used to re-weight the kernels to obtain
$\tilde{{Z}}$ through the below operations:
\begin{equation}
    \tilde{{Z}} = {Z} \odot (QK) = {Z} \odot ({Z}W{K})
\end{equation}
where $\odot$ is the Hadamard product. We show in Appendix~\ref{app:method} that the operation is equivalent to a weighted summation between each modulated frequency representation matrix, where the weights are self-generated through the queries.
We note that our proposed operation, different from \citep{rao2021global,guibas2021adaptive},
is applicable on time series with dramatic changes in input lengths and sampling rates, as we use a flexible fixed-sized multi-head filters ${K}$ that enables the transferability of the model.
Intuitively, the querying process has similarity to hypernetworks \citep{david2016hypernetworks}, which generates weights based on data itself to fully exploit the structure of the data.

Having successfully incorporated a fix-sized multi-head filter $K$ into the frequency space, 
we further explored to build nonlinearity into the operation 
through an alternative maxpooling operation $\tilde{Z} = \operatorname{MaxPool}(Z, K)$:
\begin{equation}
    \tilde{{Z}}[i,j] = \max_k |{Z}[i,j] {K}[k,j]|
\end{equation}
where the max-pooling is performed based on the absolute value of the complex features.

The resulting modulated frequency representation $\tilde{{Z}}$ is later recovered in time space through $\tilde{X} = \mathcal{F}^{-1}(\tilde{{Z}})$ with IDFT (see Figure~\ref{fig:method}(C)). We denote the whole process as $\operatorname{Freq-L}(\cdot)$, which is computationally efficient, transferrable across different input lengths and sampling rates, and can be easily implemented in a few lines of code.

\paragraph{Add $\operatorname{Freq-L}(\cdot)$ into the Transformer}
The transformer architecture has revolutionized many domains, including natural language processing \citep{devlin2018bert}, 
computer vision \citep{dosovitskiy2020image},
and recently time series processing \citep{nie2022time}.
Following \cite{nie2022time}, we first patchify the biosignals by dividing them into chunks, compute representations for each patch, and then feed the resulting patches into a transformer. Specifically, for a signal $\mathbf{s} \in \mathbb{R}^{L}$ where $L$ is the total length of the sequence, we divide them into sequences of $S = [\mathbf{s}_1, ... \mathbf{s}_N]$, where each patch $\mathbf{s}_i \in \mathbb{R}^{P}$ has a size of $P$.
An initial MLP is used to compute representation $\mathbf{x}_i = \operatorname{MLP}(\mathbf{s}_i) \in \mathbb{R}^{D}$, and the sequence is later stacked into $X_0 \in \mathbb{R}^{N\times D}$.

We replace the multi-head self-attention with our proposed multi-head frequency filter layer $\operatorname{Freq-L}(\cdot)$ to mix the information across the sequence of tokens, which gives  
the FA transformer encoder layer as below:
\begin{equation}
    X_{\ell+1}=X_{\ell}+\operatorname{Freq-L}\left(X_{\ell}\right)+\operatorname{FF}\left(X_{\ell}+\operatorname{Freq-L}\left(X_{\ell}\right)\right), \ell=\{0, \ldots, L-1\}
\end{equation}
where the representation is passed into the proposed $\operatorname{Freq-L}(\cdot)$ layer and projection layers $\mathrm{FF}(\cdot)$ with residual connections, as shown in Figure~\ref{fig:method}(C).


\subsection{Frequency-maintain pretraining with latent masking and channel independence}
\label{sec:FMmethod}

\paragraph{Masked Autoencoding in the Latent Space}
Masked autoencoder (MAE) is a self-supervised pretraining framework, which masks out input patches and predicts the missing patches using the rest present patches. 
The architecture typically contains an transformer encoder that processes non-masked patches, follows by a decoder, usually a lightweight transformer, that reconstructs the original patches \citep{he2022masked}.

To preserve the frequency information while being able to perform pretraining based on the masked autoencoding strategy, we perform \textit{masked autoencoding in the latent space}. 
Specifically, denote our frequency-aware transformer encoder as $\operatorname{FA-Enc}(\cdot)$, full sequence of biosignals $S$ is learnt through $\operatorname{FA-Enc}(\cdot)$ to obtain $X_L = [\bm{x}_1^{L}, \bm{x}_2^{L}, ..., \bm{x}_N^{L}]$. 
We sample a random set of patches based on a fixed masking ratio without replacement, 
and then process the resulting sequence with a lightweight transformer {(second)} encoder. We later pad the masked patches with mask tokens, and pass the resulting sequence into a lightweight transformer decoder to reconstruct 
the original signal, where the $i$-th reconstructed patch corresponds to $\boldsymbol{s}_i$. Denote the masked autoencoder as $\operatorname{MAE}(\cdot)$, 
\ours aims to optimize the below objective:
\begin{equation}
\label{eq:MAE}
    \mathcal{L} = \frac{1}{\Omega} \sum_{i \in \Omega} l(\boldsymbol{s}_i, \operatorname{MAE}(\operatorname{FA-Enc}(S))[i])
\end{equation}
where $i$ is the token index, $\Omega$ is the set of masked tokens, and $l$ is an error term which is 
set as mean squared error (MSE) in this work.
We show in Section~\ref{sec:results} that the performance is robust if we remove $\operatorname{MAE}(\cdot)$ and only keep $\operatorname{FA-Enc}(\cdot)$ at test time. 
We note that this is the first work that finds using the masked autoencoding objective itself, without any contrastive terms,  is effective on biosignals \citep{zhang2022self}.

\paragraph{Channel and Modality Independence}
Biosignals are multivariate time series that often face 
channel-wise and modality-wise mismatch at test time. 
To obtain robust transfer performance,
we follow previous works to use 
\textit{channel-independent design} {before the second encoder} to model multimodal biosignals \citep{liu2022seeing,nie2022time}.

Given a multi-channel biosignal $[S_1, S_2, ..., S_C]$, where $C$ denotes the total amount of channels.
We perform the channel independence learning such that each $S_{\xi}$ are passed into $\operatorname{FA-Enc(\cdot)}$ and $\operatorname{MAE}(\cdot)$ as below:
\begin{equation}
\label{eq:MAE}
    \mathcal{L} = \frac{1}{\Omega} \sum_{i \in \Omega} l(\boldsymbol{s}_i, \operatorname{MAE}([\operatorname{FA-Enc}(S_1), ...,\operatorname{FA-Enc}(S_C)])[i])
\end{equation}
where $\Omega$ is the union of masked tokens for each channels, which is independently determined based on a fixed masking ratio for each channel. 
{The parameter weights of the frequency-aware transformer encoder $\operatorname{FA-Enc(\cdot)}$ are shared across channels, creating representations that are fed into the $\operatorname{MAE}(\cdot)$, which combines information from different pretraining modalities.
}
By combining the channel independence design into our multimodal masked autoencoding objective, our architecture 
can process input signals of any channel size and order, making it robust to multimodal distributional shifts when modalities are unavailable at test time.

\section{Experiments}
\label{sec:results}

\subsection{Transfer experiments on unimodal time series}
\label{sec:transfer}

\begin{table}[t]
    \centering
    \resizebox{\linewidth}{!}{%
    \begin{tabular}{c|cccc|cccc}
        \multicolumn{9}{c}{\textit{I. Generalization with modality or task association.}} \\
         &  \multicolumn{4}{c}{\textbf{Epilepsy (EEG)}} & \multicolumn{4}{c}{\textbf{SleepEOG}}\\
        Models & Accuracy & Precision & Recall & F1 & Accuracy & Precision & Recall & F1 \\
        \hline
        TS-SD {\tiny \citep{shi2021self}}
        & 80.18 & 76.47 & 89.52 & 77.67 & 48.90 & 28.59 & 25.43 & 23.68 \\
        Mixing-up {\tiny \citep{wickstrom2022mixing}} 
        & 80.21 & 40.11 & 50.00 & 44.51 & - & - & - & -  \\
        TS2vec {\tiny \citep{yue2022ts2vec}} 
        & 93.95 & 90.59 & 90.39 & 90.45 & 67.90 & 58.23 & 62.15 & 59.28 \\
        CLOCS {\tiny \citep{kiyasseh2021clocs}} 
        & 95.07 & 93.01 & 91.27 & 92.06 & 66.86 & 56.67 & 58.99 & 57.34 \\
        TS-TCC {\tiny \citep{eldele2021time}} 
        & 92.53 & {94.51} & 81.81 & 86.33 & {69.65} & 61.56 & 61.49 & 61.16 \\
        TF-C {\tiny \citep{zhang2022self}} 
        & 94.95 & \textbf{94.56} & 89.08 & 91.49 & 69.58 & {62.04} & {68.05} & {64.15} \\
        PatchTST {\tiny \citep{nie2022time}} 
        & 95.01 & 91.66 & \textbf{92.96} & 92.27 & 68.00 & 61.20 & \textbf{68.28} & 63.26 \\
        \textbf{\ours}(scratch) & 90.41 & 84.64 & 86.29 & 85.33 & 68.29 & 60.03 & 66.10 & 61.81 \\
        \textbf{\ours}(unimodal) & \textbf{95.51} & \textbf{94.02} & 91.57 & \textbf{92.72} & \textbf{70.03} & \textbf{63.37} & 68.00 & \textbf{65.05} \\
        \hline
        \textbf{\ours}(multimodal) & \textbf{95.71} & 93.57 & \textbf{92.82} & \textbf{93.18} & \textbf{71.55} & \textbf{64.80} & \textbf{68.70} & \textbf{66.62} \\
        $\Delta$(uni, multi) & \textcolor{cyan}{$\uparrow$0.20} & \textcolor{olive}{$\downarrow$0.45} & \textcolor{cyan}{$\uparrow$1.25} & \textcolor{cyan}{$\uparrow$0.46} & \textcolor{cyan}{$\uparrow$1.52} & \textcolor{cyan}{$\uparrow$1.43} & \textcolor{cyan}{$\uparrow$0.70} & \textcolor{cyan}{$\uparrow$1.57} \\
        \hline
        \hline
        \multicolumn{9}{c}{} \\
        \multicolumn{9}{c}{\textit{II. Generalization without explicit association.}} \\
        &  \multicolumn{4}{c}{\textbf{ExpEMG}} & \multicolumn{4}{c}{\textbf{FD-B (Electromechanics)}} \\
        Models & Accuracy & Precision & Recall & F1 & Accuracy & Precision & Recall & F1 \\
        \hline
        TS-SD {\tiny \citep{shi2021self}} 
        & 46.06 & 15.45 & 33.33 & 21.11 & 55.66 & 57.10 & 60.54 & 57.03 \\
        Mixing-up {\tiny \citep{wickstrom2022mixing}} 
        & 30.24 & 10.99 & 25.83 & 15.41 & 67.89 & 71.46 & 76.13 & 72.73  \\
        TS2vec {\tiny \citep{yue2022ts2vec}} 
        & 78.54 & 80.40 & 67.85 & 67.66 & 47.90 & 43.39 & 48.42 & 43.89 \\
        CLOCS {\tiny \citep{kiyasseh2021clocs}} 
        & 69.85 & 53.06 & 53.54 & 51.39 & 49.27 & 48.24 & 58.73 & 47.46 \\
        TS-TCC {\tiny \citep{eldele2021time}} 
        & 78.89 & 58.51 & 63.10 & 59.04 & 54.99 & 52.79 & 63.96 & 54.18 \\
        TF-C {\tiny \citep{zhang2022self}} 
        & 81.71 & 72.65 & 81.59 & 76.83 & 69.38 & {75.59} & 72.02 & 74.87 \\
        PatchTST {\tiny \citep{nie2022time}} 
        & 92.68 & 90.87 & 94.51 & 92.07 & 67.03 & 71.96 & 75.57 & 70.09 \\
        \textbf{\ours}(scratch) & 93.17 & 88.58 & 94.10 & 89.97 & 67.92 & 76.45 & {76.51} & {76.20} \\
        \textbf{\ours}(unimodal) & \textbf{98.05} & \textbf{97.07} & \textbf{96.63} & \textbf{96.40} & \textbf{76.58} & \textbf{83.28} & \textbf{82.85} & \textbf{82.63} \\
        \hline
        \textbf{\ours}(multimodal) & \textbf{98.54} & \textbf{96.67} & \textbf{98.95} & \textbf{97.64} & \textbf{78.18} & \textbf{84.99} & \textbf{84.01} & \textbf{83.75} \\
        $\Delta$(uni, multi) & \textcolor{cyan}{$\uparrow$0.49} & \textcolor{olive}{$\downarrow$0.40} & \textcolor{cyan}{$\uparrow$2.32} & \textcolor{cyan}{$\uparrow$1.24} & \textcolor{cyan}{$\uparrow$1.60} & \textcolor{cyan}{$\uparrow$1.71} & \textcolor{cyan}{$\uparrow$1.16} & \textcolor{cyan}{$\uparrow$1.12} \\
        \hline
        \hline
    \end{tabular}}
    \caption{\textit{Transfer experiments on unimodal time series}.
    All benchmark models are pretrained on the same single-lead EEG. All variants of our model is based on the same architecture, where \ours (scratch) is trained from scratch, \ours (unimodal) follows the same pretraining as baselines, and \ours (multimodal) is pretrained on the multimodal version of the data. Model standard deviation are in Appendix~\ref{appsec:variation}.}
    \label{tab:single_modal}
    \vspace{-4mm}
\end{table}


\paragraph{Datasets} 
We first evaluate the model's generalization ability by transferring it on a diverse set of unimodal time series downstream tasks, following \cite{zhang2022self}. 
The transfer experiments include a set of four downstream tasks: Epilepsy \citep{andrzejak2001indications} (EEG measurement of disordered brain activity, sampling rate 174Hz with length 178); SleepEOG \citep{kemp2000analysis} (EOG measurement of each sleep stage, sampling rate 100Hz with length 3000); ExpEMG \citep{goldberger2000physiobank} (EMG measurement of muscular disorders, sampling rate 4000Hz with length 1500); FD-B \citep{lessmeier2016condition} (Electromechanical measurement of motor disorder, sampling rate 64000Hz with length 5120). We performed data pre-processing following the same protocol and data split as in \cite{zhang2022self}, more details are in Appendix~\ref{app:data}.
For model pretraining, we used the SleepEDF dataset \citep{kemp2000analysis} as in \citep{eldele2021time,zhang2022self}, where the single-channel EEG (channel Fpz-Cz) is commonly used for unimodal pretraining.
In this work, we also used an additional EEG channel (Pz-Oz) and an additional modality (EOG) from SleepEDF to perform multimodal pretraining with the same train/test split as in \cite{eldele2021time}.


\paragraph{Experimental Details}
For \ours\hspace{-1mm}, we used a 4-layer encoder, 8-head filter with 64 dimensions. The model was trained using an Adam optimizer with $\beta_1 = 0.9$, $\beta_2 = 0.99$, and a learning rate of 0.001.
{We performed a grid search based on the validation set to select the model hyperparameters (see Appendix~\ref{app:results}).}
{Following prior works, we performed full model fine-tuning on all tasks (see details in Appendix~\ref{appsec:transfer})}.
In contrast to state-of-the-art contrastive architectures \citep{eldele2021time,zhang2022self}, we did not apply data augmentation in our architecture as we found there was minimal impact on performance.  We repeated experiments with five random seeds for major results, and three random seeds for ablation experiments {(see model variation in Appendix~\ref{appsec:variation})}.
To benchmark our method, we selected an extensive set of existing state-of-the-art models, including temporal-spatial methods \citep{shi2021self,yue2022ts2vec}, contrastive methods \citep{kiyasseh2021clocs,eldele2021time}, transformers and frequency-aware approaches \citep{nie2022time,zhang2022self}. All benchmark models were pretrained on unimodal EEG under the same data split, providing a conclusive list of models for fair comparison. 

\paragraph{Pretraining on Unimodality}

Following previous works \cite{zhang2022self}, we first performed pretraining on a single-channel EEG from the SleepEDF dataset, and then fine-tuning on a small amount of data from the downstream tasks. 
The performance of our proposed architecture is shown in Table~\ref{tab:single_modal}.
We show that with the same unimodal pretraining setup on single-channel EEG, our model consistently outperforms state-of-the-art benchmarks in most experiments, 
giving $\uparrow$$4.2\%$ improvements in accuracy.
These results demonstrate that \ours is effective in terms of transfer on different tasks, with robustness to domain shifts across tasks, subjects, sampling rate, and sensors.
Surprisingly, our architecture, without any pretraining (scratch), also provides robust performance on many datasets, different from previously reported results \citep{zhang2022self}. This further demonstrates the robustness of our proposed architecture.


\paragraph{Extending Pretraining to Multimodality}

While the Fpz-Cz EEG channel is shown to be the most informative channel for the pretraining task and typically provides robust prediction performance on its own \citep{supratak2017deepsleepnet}, in this work, we explore whether using additional multimodal information from the same task can further boost the pretraining performance.
As shown in Table~\ref{tab:single_modal},
for \ours\hspace{-1mm}, including multimodal information during pretraining provides better results than unimodal pretraining in general, consistently outperforming unimodal pretraining. Training on multimodal data also improves the model's stability by giving a lower standard deviation, as shown in Appendix~\ref{app:results}. 
Note that in previous work \citep{zhang2022self}, including multimodal information hurt performance rather than helped.
This suggests that \ours can effectively utilize and combine information across modalities, resulting in better performance on downstream tasks.
We hypothesize that pretraining on multiple modalities exposes the model to a more diverse range of frequency components, improving the model's few-shot generalization.


\begin{table}
\parbox{.25\linewidth}{
\centering
\begin{tabular}{ccc}
FA & FM & Acc. \\
\hline
\xmark & \xmark & 80.68 \\
$\checkmark$ & \xmark & 84.09 \\
\xmark & $\checkmark$ & 83.53 \\
\cellcolor{gray!20}$\checkmark$ & \cellcolor{gray!20}$\checkmark$ & \cellcolor{gray!20}\textbf{85.04} \\
\hline
\end{tabular}
\caption{\small Average accuracy without FA/FM modules.}
\label{tab:FAFM}
}
\hfill
\parbox{.35\linewidth}{
\centering
\begin{tabular}{ccc}
Enc-2 & Modality & Acc. \\
\hline
\multirow{2}{*}{\xmark} & Uni & 85.04 \\
& Multi & 83.92 \\
\multirow{2}{*}{$\checkmark$} & \cellcolor{gray!20}Uni & \cellcolor{gray!20}85.05 \\
& \cellcolor{gray!20}Multi & \cellcolor{gray!20}\textbf{85.99} \\
\hline
\end{tabular}
\caption{\small The effect of keeping the 2nd encoder for multimodal pretraining.}
\label{tab:2ndEnc}
}
\hfill
\parbox{.35\linewidth}{
\centering
\begin{tabular}{ccccc}
& & \multicolumn{3}{c}{\textit{Masking ratio}} \\
\hline
& & 0.3 & 0.5 & 0.7 \\
\parbox[t]{2mm}{\multirow{3}{*}{\rotatebox[origin=c]{90}{\textit{Patch}}}}
& 10 & \cellcolor{red!10}83.86 & \cellcolor{red!10}84.05 & \cellcolor{cyan!10}82.70 \\
& 20 & \cellcolor{red!10}84.11 & \cellcolor{red!20}\textbf{85.04} & \cellcolor{red!10}83.86 \\
& 50 & \cellcolor{cyan!10}80.88 & \cellcolor{cyan!10}80.84 & \cellcolor{cyan!10}80.64 \\
\hline
\end{tabular}
\caption{\small The effect of different masking ratios and patch sizes.}
\label{tab:masking}
}
\end{table}
\paragraph{Ablations Experiments on Transferability} We performed a set of ablation experiments to understand what makes \ours robust under the transfer experiments setting {(more in Appendix~\ref{appsec:param})}. In Table~\ref{tab:FAFM}, we first studied the effect of the frequency-aware (FA) and frequency-maintain (FM) modules by either replacing the FA module with a self-attention transformer; or by replacing the FM module with a normal masking procedure. 
We found both approaches, when applied independently, improve the performance of a baseline variant by a significant margin ($\approx 3\%$). Combining both modules gives the best performance, further boosting the effect of each individual component ($\approx 5\%$).
We also tested whether it is possible to discard the second encoder at test time, which would indicate whether or not the FA encoder plays a major role in learning. 
Interestingly, we show that discarding the second encoder at test time gives almost identical performance in the unimodal setting. However, when multimodal information is used for pretraining, discarding the second encoder would give a performance that is lower than the unimodal result, while keeping the second encoder increases the unimodal performance by $\approx 1\%$ instead (see Table~\ref{tab:2ndEnc}). 
We hypothesize that it is beneficial to retain the second encoder at test time under the multimodal setting because it is responsible for merging the information present across the multimodal data.
Finally, in Table~\ref{tab:masking}, we investigate how different patch sizes and masking ratios affect the performance of our model. We show that \ours gives stable performance when the patch size is relatively small, giving robust performance under a range of masking ratios.

\subsection{Multi-modal evaluations and visualizations}

\paragraph{Datasets and Experimental Details} After verifying the model's generalization ability on transfer tasks, we investigated how well the model performs when applied to real-world cases in which multimodal information is available at test time.
To understand this, we systematically studied different combinations of the EEG Fpz-Cz, EEG Pz-Oz, EOG, EMG, and the respiration channels of the SleepEDF dataset \citep{kemp2000analysis}, which are simultaneously recorded. We followed the same train/val/test split as in \cite{eldele2021time} while attaching the multimodal information instead of using only the unimodal information.
We utilized the same model setup as in Section~\ref{sec:transfer}, aside from that we follow Section~\ref{sec:FMmethod} to expand the training and testing under multimodal designs with weight sharing and channel independence. We also implemented two variants of multimodal latent expansion methods as in Appendix~\ref{app:method}.


\begin{figure}[t] 
\begin{center}
\includegraphics[width=\textwidth]{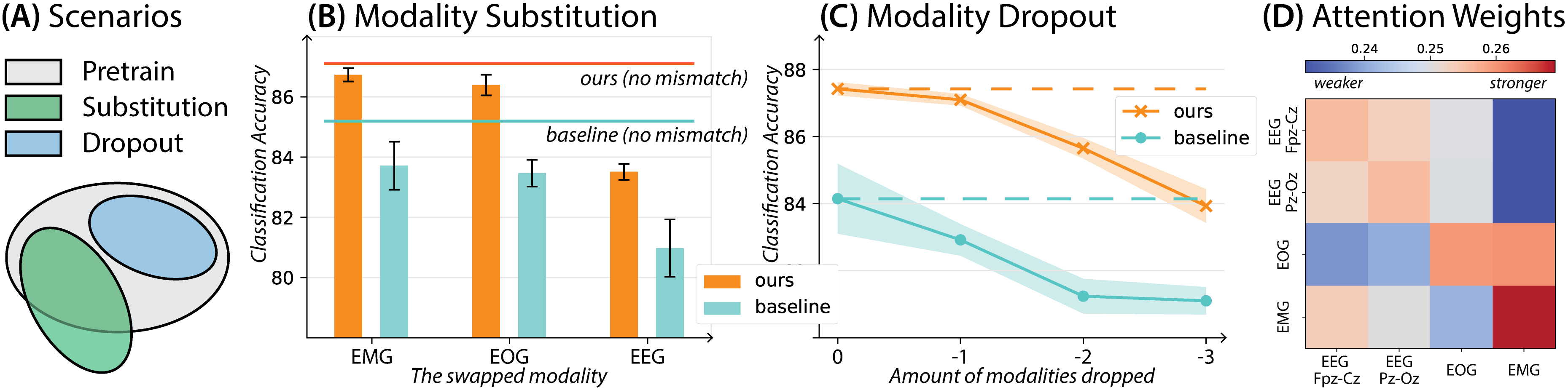}
\vspace{-7mm}
\caption{\small \textit{Multimodal evaluation results}. \textbf{(A)} Two modality mismatch scenarios are considered: Modality substitution and modality dropout. \textbf{(B)} When a modality is swapped with another available one, or \textbf{(C)} when modalities are dropped out at test time, our model gives lower performance degradation when comparing to a robust baseline. 
 \textbf{(D)} By visualizing the attention weights across modalities, we can understand how modalities are associated with each other.}      
\label{fig3}
\vspace{-4mm}
\end{center}   
\end{figure}

\paragraph{Robustness for Modality Mismatch Scenarios}
We consider two modality mismatch scenarios 
as shown in Figure~\ref{fig3}(A): (i) Modality substitution, where one modality is replaced by another modality; and (ii) Modality dropout, where only a subset of modalities is present at test time.
We show the model's performance with modality substitution in Figure~\ref{fig3}(B), where the model is pretrained with $\{$ EEG Fpz-Cz; EOG; EMG $\}$. Each of the pretraining modality is replaced with another channel to examine the performance degradation (more details in Appendix~\ref{app:multisetup}). Our model gives better performance than the robust baseline PatchTST \citep{nie2022time}, exhibiting less performance degradation. In terms of modality dropout, we pretrained the model with $\{$ EEG Fpz-Cz; EEG Pz-Oz; EOG; EMG $\}$, and we dropped an increasing amount of modalities till there is only one modality left (see Figure~\ref{fig3}(C)).
We see that \ours is more resistant to unexpected modalities dropout in comparison to the baseline.
Unlike many other baselines that contain spatial layers, \ours can be applied at test time even when there are unexpected amount of channels while exhibiting resilience 
towards modality mismatch scenarios. 
This study further demonstrated that \ours presents a robust model when used in real-world scenarios.

\paragraph{Visualizing the Connections Across Modalities} 
To understand how the information across different channels affects each other, we visualized the averaged attention matrix to examine the relationship across modalities. As shown in Figure~\ref{fig3}(D), for each channel (row), the intensity of its attention or connection to the other channels can be visualized by the color (red means stronger connections). 
Interestingly, we notice that while each channel would rely on its own information the most, they tend to focus on the stronger modalities, which is the EEG Fpz-Cz channel in our case. 
Moreover, interesting asymmetry is observed for EOG-EMG, as EOG correlates more to the EMG while the opposite does not hold. 
We hypothesize that this is because facial movement would produce moving artifacts for EOG on the temple, while the opposite connection does not hold.
This observation demonstrates that \ours 
can be used by researchers to further understand the information overlap across modalities \citep{bird2020cross}.

\vspace{-2mm}
\section{Conclusion}
\vspace{-2mm}

In this work, we proposed 
a frequency-aware masked autoencoder that performs pretraining on multimodal biosignals. Our proposed method leverages a frequency-aware encoder with fixed-size Fourier-based operator to extract representation on biosignals, and uses a frequency-maintain pretraining module to perform pretraining.
We performed extensive empirical experiments to show that (i) our model achieves state-of-the-art performance on a set of transfer experiments, where the models, both pretrained on unimodality and multimodality, can be adapted to effectively classify time series with varying input lengths, sensors, and sampling rates; and (2) our model demonstrates resilience to within-modal and across-modal distributional shifts, shows robust performance when applied in modality mismatch scenarios that are common in real-world applications.

While our model provides a good balance between utilizing frequency-information and operating on time domain, we note that, 
just like other frequency-aware architectures \citep{li2020fourier},
it remains underexplored how to interpret the specific band and type of frequency information that is taking effect in each downstream task. Exploring how the learned frequency filters can be structured and interpreted will be an exciting line of future research.
Also, in our current formulation, we only consider low-density biosignal recording systems due to the lack of publicly available high-dimensional multimodal biosignal datasets. Given the constraints, our architecture relies on the channel-independent design, which is known to suffer from capacity and robustness trade-off \citep{han2023capacity}. Extending and scaling our approach to high-dimensional sensor inputs is another exciting line of future research for modeling comprehensive human states.

\bibliography{iclr2024_conference}
\bibliographystyle{iclr2024_conference}

\newpage
\appendix
\section*{Appendix}

\section{Additional results}
\label{app:results}

\subsection{Parameter efficiency and additional ablations}
\label{appsec:param}

\paragraph{{Parameter efficiency}}

{To understand the parameter efficiency and the throughput of our approach, we compute the parameters and FLOPs between baselines and our approach in Table~\ref{tab:param}.}

\begin{table}[h]
    \centering
    \begin{tabular}{cccccc}
        \hline
        & TS2vec & TFC & TS-TCC & PatchTST & \textbf{Ours} \\
        \hline
        Params & 632K & 1.18M & 140K & 612K & 243K \\
        FLOPs & 0.69B & 1.38B & 1.95B & 35.0B & 9.42B \\
        \hline
    \end{tabular}
    \caption{{Comparison of parameters and FLOPs between baselines and our approach. The FLOPs are computed over a batch of SleepEDF data with batch size 64.}}
    \label{tab:param}
\end{table}

{We can see that, \ours is very parameter-efficient due to its fix-size frequency filter design. With the same depth (4), heads (8), and dimensionality (64), bioFAME contains only $\approx$40$\%$ parameters of the transformer baseline PatchTST. The parameter size of bioFAME also stands competitive with many CNN-based architectures.
The FLOPs of bioFAME are significantly lower than the transformer baseline PatchTST ($<$30$\%$); yet greater than CNN-based architectures.
}

\paragraph{{Additional ablations}}

{To understand the models’ sensitivity towards different hyperparameters and understand if \ours can provide better performance with increased complexity, we conducted additional ablation experiments in Table~\ref{tab:latentdim} and Table~\ref{tab:netdepth}.}

\begin{table}[h]
    \centering
    \begin{tabular}{ccccc}
        \hline
        dim & 32 & 64 & 128 & 256 \\
        \hline
        ExpEMG & 91.1 & 98.05 & 96.48 & 97.78 \\
        FD-B & 76.74 & 76.58 & 78.14 & 80.87 \\
        Avg. & 83.92 & 87.32 & 87.31 & 89.33 \\
        \hline
    \end{tabular}
    \caption{{
    Performance of our approach with different latent dimensionality.}}
    \label{tab:latentdim}
\end{table}
\begin{table}[h]
    \centering
    \begin{tabular}{ccccc}
        \hline
        depth & 3 & 4 & 5 & 6 \\
        \hline
        ExpEMG & 77.54 & 76.58 & 76.79 & 78.99 \\
        FD-B & 97.78 & 98.05 & 95.55 & 92.59 \\
        Avg. & 87.66 & 87.32 & 86.17 & 85.79 \\
        \hline
    \end{tabular}
    \caption{{
    Performance of our approach with different encoder depth.}}
    \label{tab:netdepth}
\end{table}

{We observed that increasing the latent dimensionality could further improve the performance of our approach; while increasing the network depth gives no performance gains.
}

\subsection{Data efficiency and operator selection}

\paragraph{Data efficiency}

\begin{figure}[h] 
\begin{center}
\includegraphics[width=0.8\textwidth]{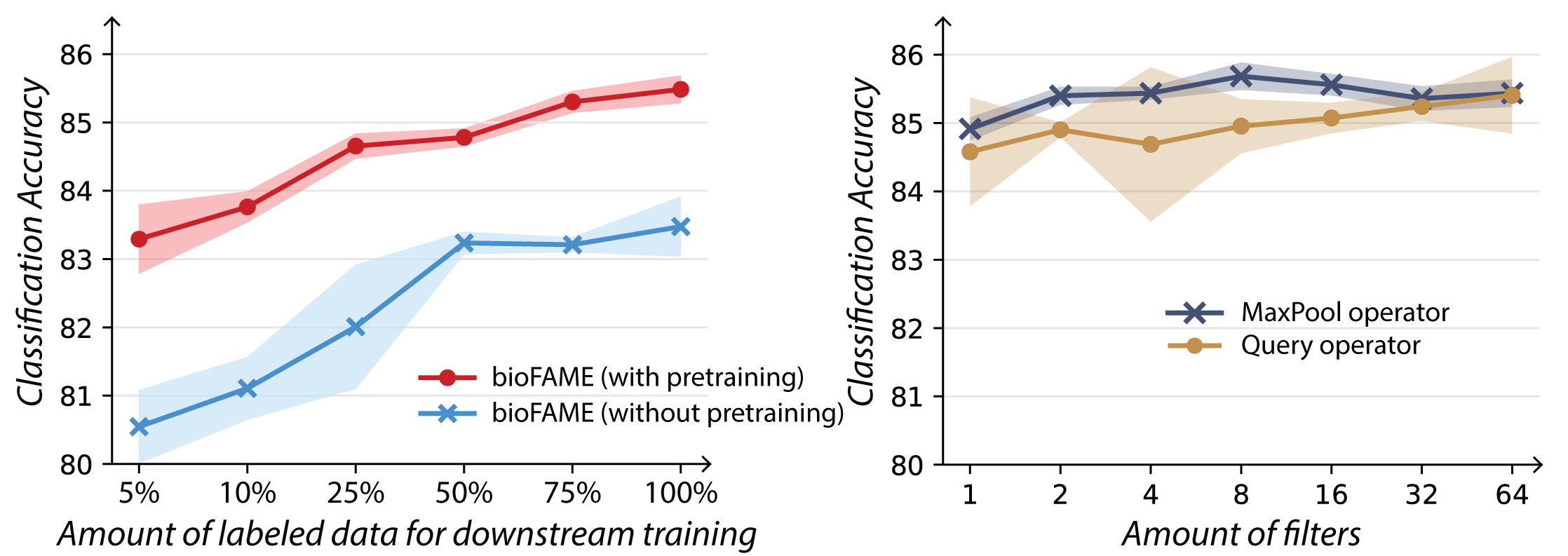}
\caption{\small \textbf{(A)} We examine the performance of \ours under low-data regime with and without pretraining. \textbf{(B)} We examine how the MaxPool operator and Query operator would perform with different amounts of filters.}
\label{fig:app-add}
\end{center}
\end{figure}

To understand the behavior of \ours within the context of limited data availability,
we conducted experiments aimed at gauging the architecture's efficacy when exposed to a reduced amount of labeled data during the finetuning phase.
We show the performance of \ours in Figure~\ref{fig:app-add}(A), both with and without pretraining, where the performance of \ours is plotted when the amount of labeled data for downstream training varies from $5\%$ to $100\%$. Notably, in contrast to previous work \citep{eldele2021time}, wherein architecture performance substantially deteriorated with decreased labeled data,
\ours achieves stable results with relatively low decay of performance even without pretraining.
Furthermore, the pretrained version of \ours gives consistently robust performance across the spectrum of labeled data proportions. 
We hypothesize that modeling biosignals using the Fourier function group with frequency operators improves the data efficiency of models.

\paragraph{Ablations on the two operators}

To validate the effectiveness of the Maxpool operator and the Query operator as described in Section~\ref{sec:methodFA}, we examine the model's performance by varying the number of filters. We find that the Maxpool operator gives more stable results, while the Query operator seems to scale better to larger amount of filters.

\subsection{Model variation}
\label{appsec:variation}

For the transfer experiments result as shown in Table~\ref{tab:single_modal}, we provide the standard variation across five different random seeds in Table~\ref{tab:var}.
Note that the entire training process, both the pretraining and the finetuning stages, is repeated to obtain the standard variation for fair evaluation. 
We notice that multimodal pretraining typically gives a lower standard deviation than that of unimodal pretraining, demonstrating that multimodal pretraining might help with the stability of the model, as it is exposed to a variety of frequency components.

\begin{table}[h]
    \centering
    \resizebox{\linewidth}{!}{%
    \begin{tabular}{c|cccc|cccc}
         &  \multicolumn{4}{c}{\textbf{Epilepsy (EEG)}} & \multicolumn{4}{c}{\textbf{SleepEOG}}\\
        Models & Accuracy & Precision & Recall & F1 & Accuracy & Precision & Recall & F1 \\
        \hline
        \textbf{\ours}(scratch) 
        & 1.17 & 2.42 & 0.72 & 1.26
        & 0.77 & 0.67 & 0.50 & 0.76 \\
        \textbf{\ours}(unimodal) 
        & 0.35 & 0.37 & 1.17 & 0.65
        & 1.39 & 1.23 & 0.91 & 0.61 \\
        \hline
        \textbf{\ours}(multimodal) 
        & 0.17 & 0.51 & 0.21 & 0.24
        & 0.90 & 0.79 & 0.89 & 0.88 \\
        $\Delta$(uni, multi) & \textcolor{cyan}{$\downarrow$0.18} & \textcolor{olive}{$\uparrow$0.14} & \textcolor{cyan}{$\downarrow$0.96} & \textcolor{cyan}{$\downarrow$0.41} & \textcolor{cyan}{$\downarrow$0.49} & \textcolor{cyan}{$\downarrow$0.44} & \textcolor{cyan}{$\downarrow$0.02} & \textcolor{olive}{$\uparrow$0.27} \\
        \hline
        \hline
        \multicolumn{9}{c}{} \\
        &  \multicolumn{4}{c}{\textbf{ExpEMG}} & \multicolumn{4}{c}{\textbf{FD-B (Electromechanics)}} \\
        Models & Accuracy & Precision & Recall & F1 & Accuracy & Precision & Recall & F1 \\
        \hline
        \textbf{\ours}(scratch) 
        & 2.67 & 3.13 & 2.25 & 3.15
        & 1.63 & 1.33 & 1.20 & 1.09 \\
        \textbf{\ours}(unimodal) 
        & 2.04 & 2.80 & 5.64 & 4.15
        & 2.74 & 1.75 & 2.01 & 2.14 \\
        \hline
        \textbf{\ours}(multimodal) 
        & 1.34 & 3.04 & 0.96 & 2.15
        & 1.94 & 1.53 & 1.44 & 1.66 \\
        $\Delta$(uni, multi) & \textcolor{cyan}{$\downarrow$0.70} & \textcolor{olive}{$\uparrow$0.24} & \textcolor{cyan}{$\downarrow$4.68} & \textcolor{cyan}{$\downarrow$2.00} & \textcolor{cyan}{$\downarrow$0.80} & \textcolor{cyan}{$\downarrow$0.22} & \textcolor{cyan}{$\downarrow$0.57} & \textcolor{cyan}{$\downarrow$0.48} \\
        \hline
        \hline
    \end{tabular}}
    \caption{The standard deviation of \ours for each transfer experiment.}
    \label{tab:var}
\end{table}

{While we believe that our diverse experiments across many datasets demonstrate the robustness of our approach under randomness, we believe that another important source of randomness comes from the data split, which is fixed in this work.
}

\subsection{Ablation results breakdown}

In Table~\ref{tab:breakdown}, we report the breakdown details for the average accuracy presented in Table~\ref{tab:FAFM} and Table~\ref{tab:2ndEnc}. Our model provides robust performance across different downstream tasks consistently.

\begin{table}[h]
    \centering
    \begin{tabular}{c|c|cccc}
        & Ablations & Epilepsy & SleepEOG & ExpEMG & FD-B \\
        \hline
        \multirow{4}{*}{Table~\ref{tab:FAFM}}
        & FA\xmark \hspace{1mm} FM\xmark \hspace{1mm} & 95.01 & 68.00 & 92.68 & 67.03 \\
        & FA$\checkmark$ FM\xmark \hspace{1mm} & 95.03 & 69.73 & 98.37 & 73.23 \\
        & FA\xmark \hspace{1mm} FM$\checkmark$ & 94.81 & 68.41 & 95.94 & 74.97 \\
        & FA$\checkmark$ FM$\checkmark$ & 95.51 & 70.03 & 98.05 & 76.58 \\
        \hline
        \multirow{4}{*}{Table~\ref{tab:2ndEnc}}
        & Uni, Enc-2\xmark & 95.91 & 70.17 & 95.94 & 78.16 \\
        & Multi, Enc-2\xmark & 95.26 & 71.04 & 96.10 & 73.28 \\
        & Uni, Enc-2$\checkmark$ & 95.51 & 70.03 & 98.05 & 76.58 \\
        & Multi, Enc-2$\checkmark$ & 95.71 & 71.55 & 98.54 & 78.18 \\
        \hline
        \hline
    \end{tabular}
    \caption{Breakdown of model performance on different downstream tasks.}
    \label{tab:breakdown}
\end{table}

\section{Experimental details}

\subsection{Datasets details}
\label{app:data}

We provide additional details about the datasets we used as follows.

\paragraph{SleepEDF} The entire SleepEDF dataset contains 197 recordings of whole-night sleep, where the dataset contains 2-lead EEG, EOG, chin EMG, respiration rates, body temperature, and event markers. 
We selected a subset of the dataset from the Cassette Study following \cite{eldele2021time}, where the dataset is used to study the age effects on sleep in healthy Caucasians.
We further followed the same train/validate/test split, and removed data with incomplete modalities.
The recordings are segmented into 30 seconds of sleep for training, where each sample is associated with one of the five sleeping patterns/stages: Wake (W), Non-rapid eye movement (N1, N2, N3), and Rapid Eye Movement (REM).


\paragraph{Epilepsy}  The Epilepsy dataset contains single-lead EEG measurements from 500 subjects, where the brain activities are recorded for subjects with seizure. The classification task is based on if the subject is having a seizure episode during the recording session.


\paragraph{SleepEOG} The SleepEOG dataset is a subset of the SleepEDF dataset under the Telemetry Study, where subjects are reported to have mild difficulty falling asleep, and thus intake either temazepam or placebo before sleep. 
The EOG channel is used for classification.

\paragraph{ExpEMG} The ExpEMG dataset consists of single-channel EMG recordings from the tibialis anterior muscle of three healthy volunteers, where they (1) do not have history of neuromuscular disease; (2) suffer from chronic low back pain and neuropathy; and (3) suffer from myopathy due to longstanding history of polymyositis. The classification task aims to classify different conditions (subjects).


\paragraph{FD-B} The FD-B dataset is an electromechanical dataset, where the motor currents and vibration signals of healthy or damaged motors are recorded. The classification task aims to detect different faulty conditions of the motors based on their behavior. We found that the motor movement follows a similar frequency assumption as biosignals \citep{hooge1981experimental}, and thus used this electromechanical dataset to provide additional validation of the transfer performance of our model.

\begin{table}[h]
    \centering
    \begin{tabular}{c|ccccc}
        Datasets & Train & Validate & Test & Sampling rate & Length \\
        \hline
        Epilepsy & 60 & 20 & 11420 & 174 & 178 \\
        SleepEOG & 1000 & 1000 & 37244 & 100 & 3000 \\
        ExpEMG & 122 & 41 & 41 & 4000 & 1500 \\
        FD-B & 60 & 21 & 13559 & 64000 & 5120 \\
        \hline
    \end{tabular}
    \caption{Dataset split details for different downstream tasks.}
    \label{tab:dataset}
\end{table}


We performed the transfer experiments based on the same settings as in \cite{zhang2022self}, where we used the train/validate/test spilt as shown in Table~\ref{tab:dataset} for downstream fine-tuning to demonstrate the few-shot generalization ability of the model across different signals.

\subsection{{Model training and transfer experiments details}}
\label{appsec:transfer}

{
For all experiments, we pretrain \ours for 200 epochs on the SleepEDF dataset using a batch size of 128 to obtain the weights of the model.
During fine-tuning, we remove the lightweight second encoder that mixes information across modalities, and use the average token of the frequency-aware transformer encoder to perform the prediction for downstream tasks.
We fine-tune \ours for 80 epochs with a batch size of 64, using an Adam optimizer with a learning rate of $0.001$ on all datasets to obtain the final results. 
We perform all transfer experiments under the same training setup for all downstream tasks without additional adjustment for each dataset.
Note that we perform full-scale model finetuning instead of linear probing when performing the transfer experiments, because the former approach is shown to be more effective for transformers in previous works \citep{he2022masked}.
}

\subsection{Multimodal setup details}
\label{app:multisetup}



The multimodal experiments are designed to tackle the challenge presented by modality mismatch scenarios, where discrepancies in biosignal recording setups between training and testing phases lead to distributional shifts.
Due to the scarcity of comprehensive multimodal datasets encompassing simultaneous recording of diverse modalities of biosignals,
we exclusively used the SleepEDF dataset due to its modality coverage. 

We first empirically assessed the representation quality of each individual channel. Similar to the findings in \cite{supratak2017deepsleepnet}, we found that the representation capacity of different channels follows EEG Fpz-Cz $>$ EEG Pz-Oz $>$ EOG $>$ EMG $>$ resp. Building upon these insights, we performed the modality substitution and modality dropout experiments following the below pretraining and finetuning setup.

\begin{table}[h]
\centering
\begin{tabular}{cc}
Training modalities & Testing modalities \\
\hline
\multirow{3}{*}{\shortstack{EEG Fpz-Cz;\\ EOG; EMG}} 
& EEG Fpz-Cz; EOG; resp \\
& EEG Fpz-Cz; EEG Pz-Oz; EMG \\
& EEG Pz-Oz; EOG; EMG \\
\hline
\end{tabular}
\vspace{-2mm}
\caption{\small Modality setup for modality substitution experiments.}
\label{tab:modsub}
\end{table}

\begin{table}[h]
\centering
\begin{tabular}{cc}
Training modalities & Testing modalities \\
\hline
\multirow{3}{*}{\shortstack{EEG Fpz-Cz;\\ EEG Pz-Oz; EOG; EMG}} 
& EEG Fpz-Cz; EEG Pz-Oz; EOG \\
& EEG Fpz-Cz; EEG Pz-Oz \\
& EEG Fpz-Cz \\
\hline
\end{tabular}
\vspace{-2mm}
\caption{\small Modality setup for modality dropout experiments.}
\label{tab:masking-supp}
\vspace{-4mm}
\end{table}

\subsection{Hyperparameter searching details}
\label{app:results}

For transfer experiments, we performed hyperparameter searching based on results on the Epilepsy dataset, and used the same parameter setting across all transfer experiments. Specifically, we performed a grid search of learning rate of [0.0001, 0.001, 0.01], transformer depth of [2, 3, 4, 5, 6], latent dimensionality of [16, 32, 64, 128], dropout rate of [0.2, 0.3, 0.4], operator type, and filter amount correspondingly.
{
We followed the convention for transformers and selected the MLP dimension of 128 and head dimension of 16 for \ours and the baseline transformer.
We selected the optimal patch size and masking ratio based on results in Table~\ref{tab:masking}.
We did not search for the optimal batch size, or investigate the effect of using different activation functions or normalization techniques.}
For multimodal experiments, we evaluate the model's performance on the pretraining dataset, and performed the evaluation on the finetuning modalities using the best model used in pretraining. For the multimodal experiments, we performed a smaller scale grid search for the latent dimensionality and transformer depth.

\section{Methodology details}
\label{app:method}

\subsection{Additional explanation of motivation}


Biosignals are often analyzed in their frequency space, where they are either studied through predefined frequency regions or through aperiodic components which typically form a 1/f-like distribution \citep{donoghue2020parameterizing}.
The significance of frequency information is well-documented due to its intricate interrelation with various facets of learning, aging, as well as diseases such as ADHD or seizures.
Correspondingly, modeling approaches that rely on the manual extraction and preprocessing of spectrogram features have demonstrated robust empirical performance \citep{supratak2017deepsleepnet}.
Building upon these insights, we hypothesize that modeling biosignals employing function groups within the frequency domain could benefit the learning process by enhancing model adaptability and data efficiency.
We note that this hypothesis might be violated if the frequency components carry limited information in other formats of time series datasets.

\subsection{Intuition for the multi-head frequency filter layer}

We provide additional intuition for the design of our multi-head frequency filter layer by breaking down the computation for each individual filter.
For each $k$-th filter $K[k]$ inside $K \in \mathbb{C}^{H \times D}$, given latent representation $Z = [\bm{z}_1, \bm{z}_2, ..., \bm{z}_N]^T \in \mathbb{C}^{N \times D}$, we compute $Z^{(k)} = [\bm{z}_1 \odot K[k], \bm{z}_2 \odot K[k], ..., \bm{z}_N \odot K[k]]^T$, where $\odot$ represents the Hadamard product between each representation and the learnable filter weights.
In order to learn the combination between different filters, we define weights $\bm{w}$ that compute $\tilde{Z} = \sum_{k=1}^{H} w_k Z^{(k)}$.

To increase the expressiveness of the filtering operation, instead of learning a linear combination of different filters, we borrow intuition from the computation of self-attention to compute the queries for the kernel weights $\bm{w}$ through ${\bm{w}} = \bm{z} W$, where $W \in \mathbb{C}^{D \times H}$. Thus, we have:
\begin{equation}
\begin{aligned}
    \tilde{Z}[i, j] & = \sum_{k=1}^{H} (\sum_{j=1}^{D}Z[i, j] W[j, k]) Z[i, j] K[k,j] \\
    & = Z[i, j] \sum_{k=1}^{H} (\sum_{j=1}^{D}Z[i, j] W[j, k]) K[k,j]
\end{aligned}
\end{equation}
which gives $\tilde{Z} = Z \odot (Z W K)$. In our implementation, we use the real values of latents to learn the weights of the combiner though ${\bm{w}} = \bm{z}_{\operatorname{real}} W$.
Similarly, based on the same intuition of combining filtered matrices, we have the max pooling operation.

\subsection{Model variants for combining multimodal representations}

In transfer experiments, we use the average of tokens to extract the final representations for the downstream classification. 
However, when having multimodal information, 
fixing the dimensionality of the latent representation when many modalities are present might narrow down the information from each modality, which might cause information loss.
Thus, in multimodal experiments, we first average the representations from each individual modality, and then concat the representations across modalities before performing the downstream classification.

\end{document}